\documentclass{article}
\pdfoutput=1
\usepackage{spconf,amsmath,graphicx}
\usepackage{amssymb,amsfonts}
\usepackage{dsfont}
\usepackage{bbm}
\DeclareMathOperator*{\argmin}{arg\,min}
\usepackage{gensymb}
\usepackage{algorithm}
\usepackage{algpseudocode}
\algrenewcommand\algorithmicrequire{\textbf{Input:}}
\algrenewcommand\algorithmicensure{\textbf{Output:}}
\usepackage{graphicx}
\usepackage{textcomp}
\usepackage{booktabs}
\usepackage{subcaption}
\usepackage{xcolor}
\definecolor{onlineBlue}{RGB}{172, 197, 255}
\definecolor{onlineRed}{RGB}{255, 177, 177}
\usepackage[hidelinks]{hyperref}
\hypersetup{
    colorlinks=false,
    linkcolor=black,
    filecolor=black,
    citecolor=black,
    urlcolor=black,
}
\usepackage{enumitem}
\setlist{nosep, leftmargin=14pt}

\usepackage{mwe} 

\title{Self-Taught Semi-Supervised Anomaly Detection on Upper Limb X-rays}
\name{Antoine Spahr\textsuperscript{1}, Behzad Bozorgtabar\textsuperscript{1,2}, Jean-Philippe Thiran\textsuperscript{1,2}}
\address{\textsuperscript{1} Signal Processing Laboratory (LTS5), \'Ecole Polytechnique F\'ed\'erale de Lausanne (EPFL), Switzerland\\
\textsuperscript{2} CIBM Center for Biomedical Imaging, Switzerland}

\begin{document}
%
\maketitle
\begin{abstract}
Detecting anomalies in musculoskeletal radiographs is of paramount importance for large-scale screening in the radiology workflow. Supervised deep networks take for granted a large number of annotations by radiologists, which is often prohibitively very time-consuming to acquire. Moreover, supervised systems are tailored to closed set scenarios, e.g., trained models suffer from overfitting to previously seen rare anomalies at training. Instead, our approach's rationale is to use task agnostic pretext tasks to leverage unlabeled data based on a cross-sample similarity measure. Besides, we formulate a complex distribution of data from normal class within our framework to avoid a potential bias on the side of anomalies. Through extensive experiments, we show that our method outperforms baselines across unsupervised and self-supervised anomaly detection settings on a real-world medical dataset, the MURA dataset. We also provide rich ablation studies to analyze each training stage's effect and loss terms on the final performance.
\end{abstract}

\begin{keywords}
Anomaly detection, contrastive learning, pretext task, semi-supervised model, X-ray.
\end{keywords}
\section{Introduction}
\label{sec:intro}
X-ray is a common medical support for multiple diagnosis such as respiratory disease or bone malformation and fracture. However, examining radiographs and reporting work for the signs of a specific anomaly are time-consuming and require qualified experts. Therefore, developing a computer-aided diagnosis system to search for anomalies has become attractive in X-ray imaging. It can automate the screening and skew the allocation of resources towards the most enlightening input data samples with the diagnostic contents \cite{schlegl_unsupervised_2017}.

Contrary to classical approaches, such as one class-SVM \cite{scholkopf_estimating_2001} or kernel density estimation \cite{parzen_estimation_1962}, which often fail for high-dimensional inputs e.g., images without proper feature engineering, deep learning-based anomaly detection (AD) methods \cite{varma_automated_2019} can facilitate detection of anomalies in X-ray images more accurately. 
\begin{figure}[!t]
    \centerline{\includegraphics[width=\columnwidth]{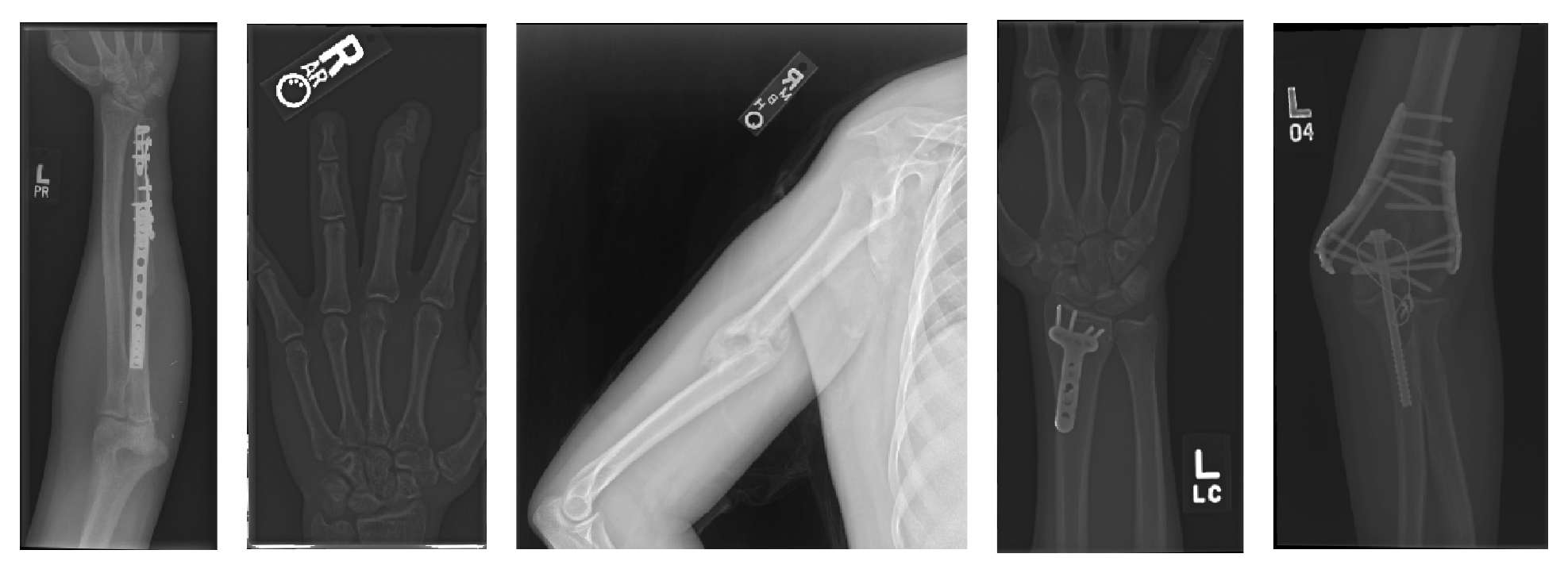}}
    \caption{Abnormal X-ray samples of the MURA dataset. \emph{Left}: Operative plate on ulna. \emph{Center Left}: Phalanx displacement. \emph{Center}: Humerus fracture. \emph{Cetner Right}: Operative plate on wrist. \emph{Right}: Operative plate on elbow.}
    \label{fig:abnormal_sample}
  \end{figure}
 A straightforward approach to the anomaly detection problem could be training a supervised classification model using both labeled normal and abnormal samples \cite{varma_automated_2019,rajpurkar_mura_2018}. However, supervised learning-based anomaly detection has several drawbacks that render its application challenging for real-world scenarios. First, such an approach requires access to a large number of labeled samples of both the normal and abnormal classes, which is often prohibitively expensive to acquire.  Second, supervised models suffer from severe class imbalance issues, and long-tailed distributions as anomalies are rare and trained models heavily biased toward negative samples. Third, another concern when deploying these supervised models is that they often fail to generalize to new domains due to a domain gap caused by different imaging scenarios. These models intend to overfit to the training anomalies, which cannot generalize well to unseen test anomalies. Finally, current supervised deep learning solutions lack interpretation, which hinders their application in an area as sensitive as final diagnosis and treatment.
 
 A more meaningful strategy involves modeling the distribution of normal samples and detecting any abnormal samples as those not falling on normal samples' distribution. This general approach copes with the challenges of determining the appropriate distribution for normal data. Indeed, there is no need to label all possible anomalies, and the normal distribution can be modeled with no or few labeled samples. Therefore, unsupervised and semi-supervised approaches are usually preferred, allowing a notable reduction of annotations and time spent on radiologists' reports. Some examples of abnormal X-rays are presented in Fig. \ref{fig:abnormal_sample}.
 
\subsection{Contributions}
  \label{subsec:contributions}
  Our contributions are as follows:
  \begin{itemize}
      \item We propose self-taught anomaly detection, a universal semi-supervised anomaly detection that can generalize well to detect unseen anomalies without using any prior domain knowledge,
      
      \item We show that our method yields better classification performances on a real-world medical dataset when there are few labeled data. This is more practical than the previous supervised and semi-supervised approaches on the assumption that there is access to a large amount of annotated data,

      \item Our proposed framework method enables faster optimization and simplifies the network architecture at inference time. Therefore, it can be easily incorporated into existing disease screening pipelines. 

  \end{itemize}
 
\section{Related Work}
  \label{sec:relatedwork}
  In this section, we provide an overview of the previous studies on anomaly detection task.
  
  Unsupervised learning-based methods \cite{zhai2016deep,bb2010comparison,bb2011genetic} take advantage of unlabeled data and can thus be adapted for the anomaly detection task. Several deep anomaly detection methods use the deep convolutional autoencoder (AE) \cite{chen_outlier_2017,sakurada_anomaly_2014}. Some recent approaches (e.g., AnoGAN \cite{schlegl_unsupervised_2017}, OCGAN \cite{tang_abnormal_2019}) built upon generative adversarial networks (GANs) \cite{rad2020benefiting,bozorgtabar2019learn,bozorgtabar2019syndemo,bozorgtabar2019informative,bozorgtabar2019using,bozorgtabar2020exprada,mahapatra2019progressive,mahapatra2020pathological,mahapatra2017image} and have been re-used to formulate anomaly detection. The underlying rationale behind those methods is to model the distribution of normal samples in an unsupervised manner, and then this knowledge is exploited to detect deviations (outliers). For example, AE-based models are usually trained by minimizing the reconstruction error on the single-class normal samples and then using the reconstruction error as the anomaly score. The occurrence of anomalies is indicated by a high reconstruction error using a predetermined threshold. However, this assumption, sometimes violated, as AE can reconstruct anomalous samples well, yielding miss detection of anomalies at test time.
  
  Some recent methods \cite{ruff_deep_nodate, ruff_deep_2019, ghafoori2020deep} proposed deep models based on support vector data description to learn a representation of normal samples that enclose them in minimum volume hyper-spheres while mapping abnormal ones outside hyper-spheres. However, their approach relies on AE pre-training to initialize the encoder's weights. This pre-training step tends to ignore the images' underlying structure as the pre-trained weights are sensitive to biased low-level features, e.g., color information.
  
  Thus, an anomaly detection task's challenge is to get the best representation of the normal data to segregate abnormal samples. Learning a representation without using labels is also commonly known as self-supervised learning \cite{jing_self-supervised_2020,kolesnikov_revisiting_2019,bb2020,zhou2020comparing}. This approach uses a \emph{pretext} task for which the labels are derived from the unlabeled data itself. To solve the pretext tasks, the datasets do not need to be manually labeled by qualified experts; instead, they can, e.g., be labeled by exploiting the relations between different input images. To do so, the deep model has to learn a meaningful representation of the data in a lower-dimensional space. For example, self-supervised deep methods \cite{gidaris2018unsupervised, golan2018deep} have been proposed to train a classifier for which a self-labeled multi-class dataset is created by applying a set of geometric transformations to the images. Contrastive learning is a special case of self-supervised learning in which the pretext task is to compare images and find the corresponding one among a set of images \cite{baldi_contrastive_1991,oord_representation_2019}. Recently, self-supervised contrastive learning has shown promising results for the image classification task \cite{he_momentum_2020, chen_simple_2020,chen_improved_2020,zhou2020comparing}. However, most of these methods are domain-specific, and their efficacy for anomaly detection has not been explored yet. 
  

We propose self-taught anomaly detection, a universal semi-supervised anomaly detection that can generalize well to detect unseen anomalies without using any prior domain knowledge. Our pre-training stage consists of two attractive and repulsive forces. The former brings the representations of similar samples closer, while the latter pushes samples of different radiographic study types apart from each other. Our proposed framework method enables faster optimization and simplifies the network architecture at inference time. Therefore, it can be easily incorporated into existing disease screening pipelines. We show that the proposed method yields a better initialization and better accuracy performance on real-world upper-limb X-rays. 
    
\section{Proposed Method}
\label{sec:CDMSAD}

    \begin{figure}[t]
        \centering
        \centerline{\includegraphics[width=8.5cm]{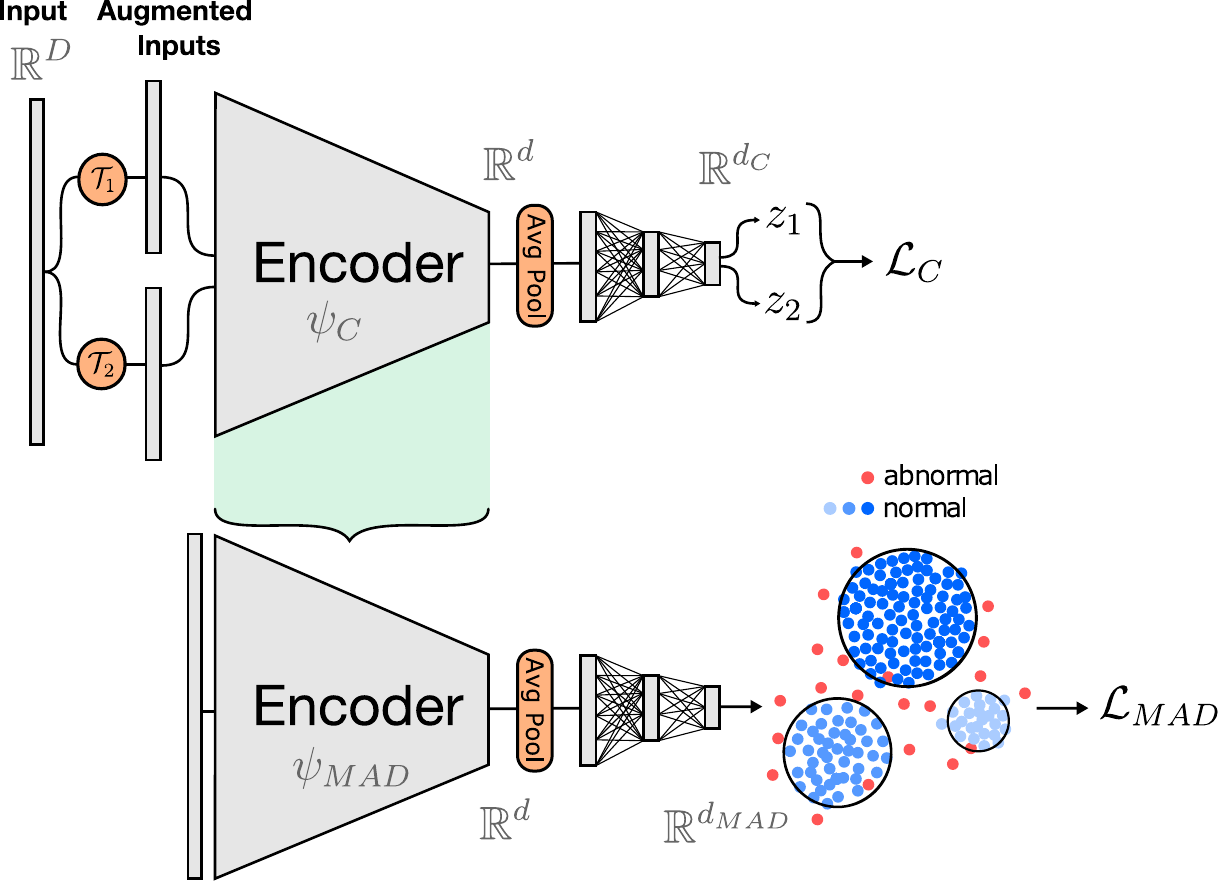}}
        \caption{The proposed training scheme. \textit{Top plot}: the network trained on the InfoNCE loss. \textit{Green bracket}: adaptation of the encoder $\psi_C$ to the anomaly detection encoder $\psi_{MAD}$. \textit{Bottom plot}: network trained on the semi-supervised multi-modes anomaly detection loss.}
        \label{fig:CDMSAD_scheme}
    \end{figure}
    
    \begin{figure}[!t]
        \begin{algorithm}[H]
          \caption{Optimization}
          \label{algo:CDMSAD}
          \begin{algorithmic}[1]
            \Require Unlabeled data $\mathbf{x}_1 ... \mathbf{x}_n$
                     \Statex \hspace{12pt} Labeled data $(\mathbf{\tilde{x}}_1,\tilde{y}_1),...,(\mathbf{\tilde{x}}_m,\tilde{y}_m)$
                     \Statex \hspace{12pt} Hyper-parameters $\eta, \gamma, N_s, \tau, \lambda$
                     \Statex \hspace{12pt} SGD learning rate $\epsilon$
            \Ensure  Trained neural network $\mathcal{W_{MAD}}^*$
            \Statex
            \ForAll{epoch}\Comment{Pre-train with pretext task}
              \ForAll{batch $\left \{ \mathbf{x}_{b} \right \}_{k=1}^{N}$}
                \State $\mathcal{L}_C \gets InfoNCE(\left \{ \mathbf{x}_{b} \right \}_{k=1}^{N}, \phi_C(\left \{ \mathbf{x}_{b} \right \}_{k=1}^{N};\mathcal{W}_C))$
                \State $\mathcal{W}_C \gets \mathcal{W}_C-\epsilon\cdot\nabla_{\mathcal{W}_C}\mathcal{L}_C$
              \EndFor
            \EndFor
            \Statex
            \State $\mathcal{W_{\psi_{MAD}}} \gets \mathcal{W}_{\psi_C}$\Comment{Encoder's weights transfer}
            \State $\mathbf{c} \gets KMeans(N_s, \phi_{MAD}(\mathbf{x};\mathcal{W_{MAD}}))$\Comment{Initialization}
            \Statex
            \ForAll{epoch}\Comment{Fine-tuning}
              \ForAll{batch $\left \{ \mathbf{x}_{b} \right \}_{k=1}^{N}$}
                \State $\mathbf{c}_k \gets \argmin{\mathbf{c}_k}\left\|\phi_{MAD}(\mathbf{x}_{b,i};\mathcal{W_{MAD}})-\mathbf{c}_k\right\|$
                \State $\mathcal{L}_{MAD} \gets MAD(\phi_{MAD}(\left \{ \mathbf{x}_{b} \right \}_{k=1}^{N};\mathcal{W_{MAD}}), \mathbf{c})$
                \State $\mathcal{W_{MAD}} \gets \mathcal{W_{MAD}}-\epsilon\cdot\nabla_{\mathcal{W_{MAD}}}\mathcal{L}_{MAD}$
              \EndFor
              
              \ForAll{center $\mathbf{c}_i$}\Comment{Prune centers}
                \State $N_i \gets $ number of normal samples assigned to $\mathbf{c}_i$
                \If{$N_i<\gamma\cdot\max{N_j}$} 
                    \State remove center $\mathbf{c}_i$
                \EndIf
               \EndFor
            \EndFor
          \end{algorithmic}
        \end{algorithm}
      \end{figure}
    
    \subsection{Overview}
    Our method uses a two-step training to learn image representations of unlabeled data using a pretext task and then adapt those representations to the actual task of semi-supervised anomaly detection. The pre-training stage aims to leverage unlabeled data in a task-agnostic way using a defined pretext objective. Let $\phi_C(\cdot,\mathcal{W}_C):\mathbb{R}^D\rightarrow\mathbb{R}^{d_C}$ be an image encoder with $L_C$ hidden layers and weights $\mathcal{W}_C=\{\mathbf{W}_C^1,...,\mathbf{W}_C^{L_C}\}$. $\phi_C$ consists of a convolutional encoder $\psi_C$ that maps an input in $\mathbb{R}^D$ to a compressed representation in $\mathbb{R}^{d}$ followed by a projection head network that further compresses the input into $\mathbb{R}^{d_C}$. Let's define a transformation $\mathcal{T}:\mathbb{R}^D\rightarrow\mathbb{R}^D$ that heavily augments the input image. The goal of the pre-training (pretext task) is to optimize the weights $\mathcal{W}_C$ of $\phi_C$ such that two versions of an image augmented by $\mathcal{T}$ are brought together in representation space $\mathbb{R}^{d_C}$. In practice, for a pair $\{\mathbf{x}_i, \mathbf{x}_j\}$ in the set of positive pairs $\mathcal{P}$, the network is trained to identify $\mathbf{x}_j$ from a set of $N$ images $\{\mathbf{x}_k\}_{k\neq i}$. This is done by maximizing the cosine similarity between the representation of a pair ($\phi_C(\mathbf{x}_i)=\mathbf{z}_i$ and $\phi_C(\mathbf{x}_j)=\mathbf{z}_j$) and minimizing the cosine similarity with respect to the other samples' representations of the set. The pretext task's objective can thus be formulated as
    \begin{equation}
        \small
        \mathcal{L}_C=\min_{\mathcal{W}_C}\sum_{(i,j)\in\mathcal{P}}-log\frac{exp(sim(\mathbf{z}_i,\mathbf{z}_j)/\tau)}{\sum_{k=1}^{2N}\mathds{1}_{[k\neq i]}exp(sim(\mathbf{z}_i,\mathbf{z}_k)/\tau)}
    \end{equation}
    where $\mathbf{z}_i = \phi_C(\mathbf{x}_i;\mathcal{W}_C)$ and $sim(\mathbf{u},\mathbf{v})=\frac{\mathbf{u}^T\mathbf{v}}{\left\|\mathbf{u}\right\|\left\|\mathbf{v}\right\|}$ and
    $\mathds{1}_{k\neq i}\in\{0,1\}$ is a function giving $1$ if $k\neq i$ else $0$. $N$ denotes the number of samples within a minibatch, and $\tau$ is a hyper-parameter called the temperature. This loss is also known as the InfoNCE loss \cite{oord_representation_2019}.

    The pre-training stage is followed by the proposed semi-supervised anomaly detection as the fine-tuning that map the normal data into multiple hyper-spheres. Formally we have access to $n$ unlabeled samples $\mathbf{x}_1,\ldots,\mathbf{x}_n \in \mathcal{X}$ with $\mathcal{X}\subseteq\mathbb{R}^D$ where $D$ is the input dimension. In addition to the unlabeled samples, we have access to few $m$ labeled samples $(\mathbf{\tilde{x}}_1,\tilde{y}_1),\ldots,(\mathbf{\tilde{x}}_m,\tilde{y}_m)\in \mathcal{X} \times \mathcal{Y}$ where $\mathcal{Y}=\{-1,+1\}$. Known normal samples are labeled as $\tilde{y}=+1$ and known abnormal samples are labeled as $\tilde{y}=-1$. Let $\phi_{MAD}(\cdot,\mathcal{W_{MAD}}):\mathbb{R}^D\rightarrow\mathbb{R}^{d_{MAD}}$ be a deep neural network with $L_{MAD}$ hidden layers and weights $\mathcal{W_{MAD}}=\{\mathbf{W}_{MAD}^1,...,\mathbf{W}_{MAD}^{L_{MAD}}\}$. The goal of the downstream task is to train a deep neural network $\phi_{MAD}$ to transform the input into a lower dimension $d_{MAD}$ such that the normal samples are enclosed in $N_s$ hyper-spheres of minimum volume centered on $N_s$ defined points in $\mathbb{R}^{d_{MAD}}$: $\mathbf{c}=\{\mathbf{c}_1,...,\mathbf{c}_{N_s}\}$, while abnormal samples are mapped away from all hyper-spheres' centers. The multi-modal anomaly detection objective ($MAD$) can be thus written as: 
    
        \begin{equation}
            \small
            \label{eq:DMSAD_obj}
            \begin{split}
            \mathcal{L}_{MAD}=\min_{\mathcal{W_{MAD}}} & \frac{1}{n+m}\sum_{i=1}^n\left\|\phi_{MAD}(\mathbf{x}_i;\mathcal{W_{MAD}})-\mathbf{c}_k\right\|^2 \\
            & + \frac{\eta}{n+m}\sum_{j=1}^m(\left\|\phi_{MAD}(\mathbf{\tilde{x}}_i;\mathcal{W_{MAD}})-\mathbf{c}_k\right\|^2)^{\tilde{y}_j} \\
            & + \frac{\lambda}{2}\sum_{l=1}^L\left\|\mathbf{W_{MAD}}^l\right\|^2
            \end{split}
        \end{equation}
        where the center $k$ is assigned as the closest one \textit{i.e.}, $k=\argmin_j\left\|\phi_{MAD}(\mathbf{x}_i;\mathcal{W_{MAD}})-\mathbf{c}_j\right\|$. The first term in Eq. \ref{eq:DMSAD_obj} penalizes unlabeled points away from the closest center since we assume that most unlabeled samples come from the normal distribution. The second term in Eq. \ref{eq:DMSAD_obj} pushes known abnormal samples away from the closest center and known normal samples toward that center. Finally, the third term in Eq. \ref{eq:DMSAD_obj} imposes a regularization on the network's weights $\mathcal{W_{MAD}}$ with hyper-parameter $\lambda$. The hyper-parameter $\eta$ controls the relevance of the labeled terms in the final objective. We opt for two-phase training instead of a joint training setting due to the difference in the data processed by each phase and requiring fewer hyper-parameters (scales of loss, etc.).

    \subsection{Optimization}
    After the pre-training stage, the weights of the encoder $\psi_C$ are used to initialize the encoder $\psi_{MAD}$ for anomaly detection. The $N_s$ hyper-sphere centers are initialized using the K-means algorithm on the embedded normal samples, and then non-meaningful clusters\footnote{Non-meaningful clusters refer to those clusters in which the cluster cardinality is not large enough or include noisy samples.} are removed progressively during the optimization procedure. The encoder's weights $\mathcal{W_{MAD}}$ are updated for the anomaly detection task (Eq.\ref{eq:DMSAD_obj}) via a stochastic gradient descent (SGD) optimization until convergence. At each step, a cluster center is kept only if the cardinality of normal samples is larger than a fraction $\gamma$ of the maximum cardinality. It ensures that the model learns the best number of centers without any \textit{a priori} on the number of modes. Upon testing, the anomaly score of a sample is given by computing the distance between its embedding and the closest hyper-sphere centers: $s_{MAD}(\mathbf{x}) = \left\|\phi_{MAD}(\mathbf{x};\mathcal{W_{MAD}})-c_k\right\|$ where $k = \argmin_j\left\|\phi_{MAD}(\mathbf{x};\mathcal{W_{MAD}})-c_j\right\|$ is the closest hyper-sphere center. The optimization procedure is presented both in Fig. \ref{fig:CDMSAD_scheme} and Algorithm \ref{algo:CDMSAD}.

\section{EXPERIMENTAL RESULTS}
\label{sec:application}

    \subsection{Dataset and Preprocessing}
    \label{ssec:dataset}
\begin{figure}[!t]
        \centerline{\includegraphics[width=\columnwidth]{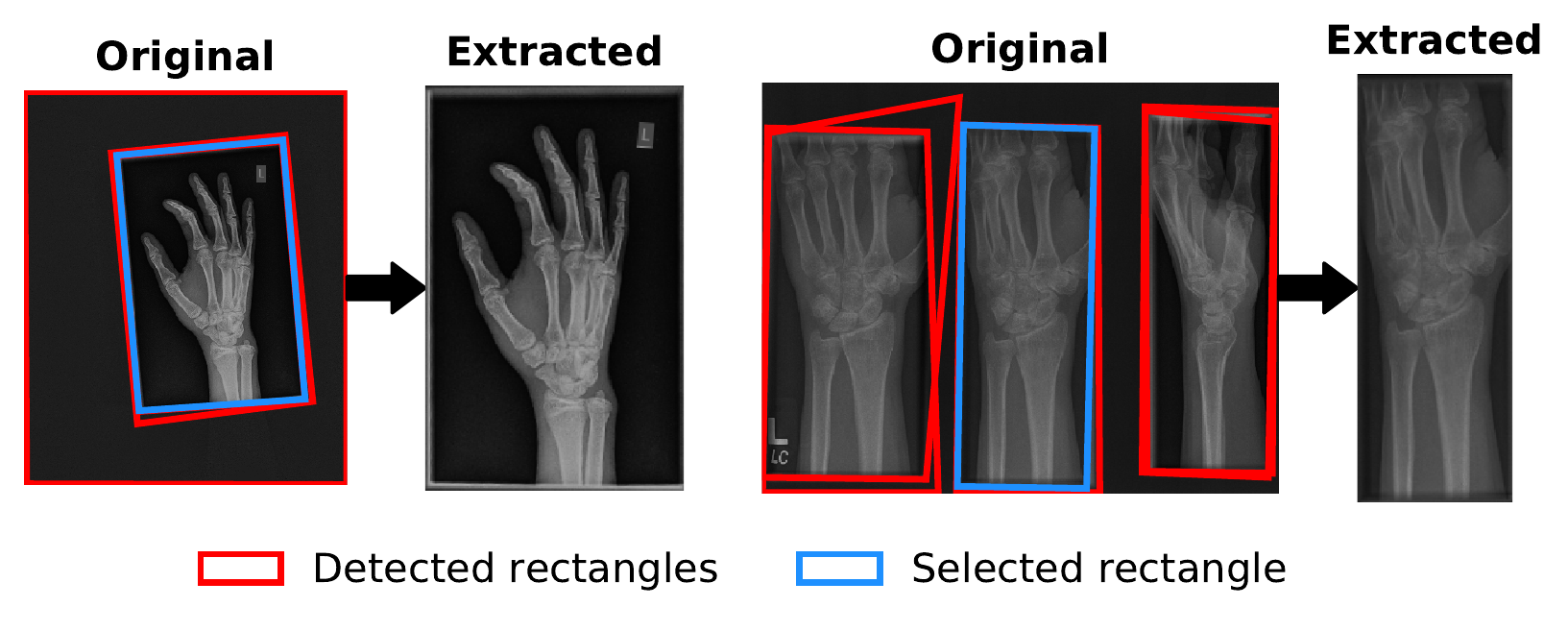}}
        \caption{Two examples of X-ray image carrier extraction for a hand and a wrist, respectively. The detected rectangles are highlighted in red, while the selected rectangle is displayed in blue.} 
        \label{fig:rect_crop}
    \end{figure}

\begin{figure}[!t]
        \centerline{\includegraphics[width=\columnwidth]{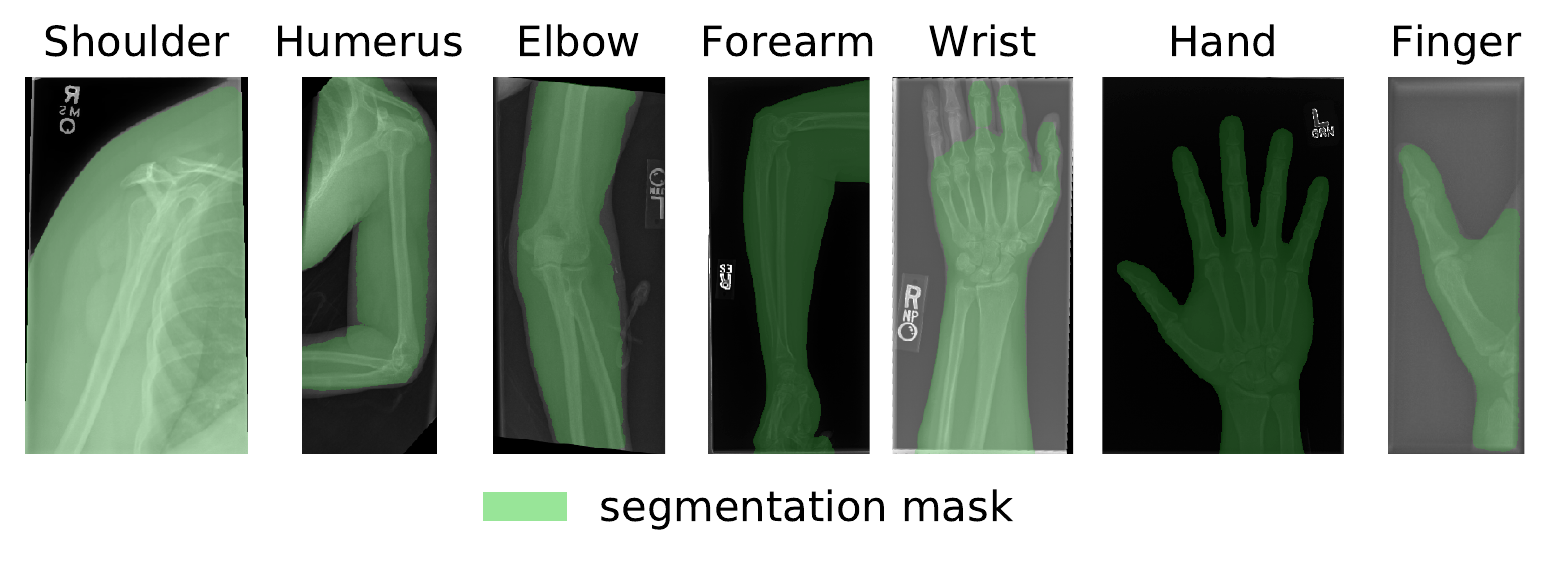}}
        \caption{Examples of segmentation results of the musculoskeletal X-rays.}
        \label{fig:segmentation}
      \end{figure}

\begin{figure*}[!t]
          \centering
          \includegraphics[width=\textwidth]{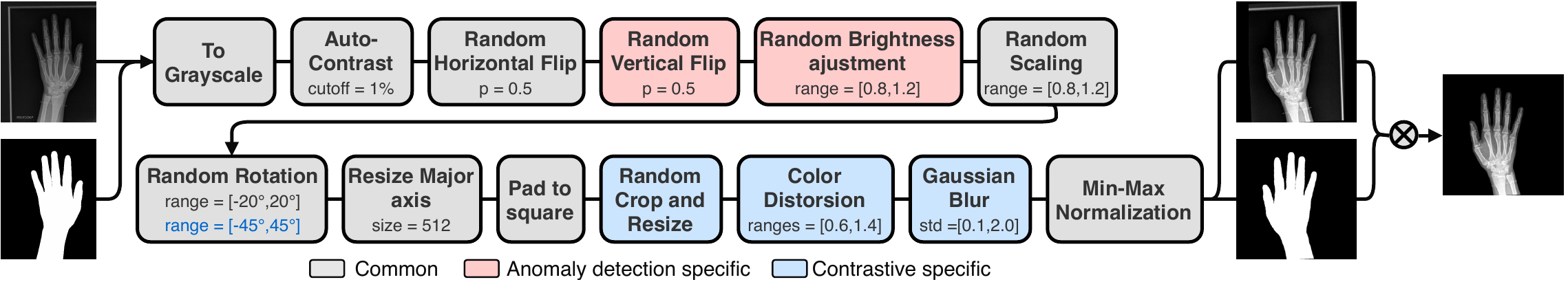}
          \caption{Online processing pipelines of both pre-training and the main training step of anomaly detection. The most relevant parameters of each transformation are presented in the corresponding transformation box.  Transformations specific to anomaly detection are highlighted in \textcolor{onlineRed}{\textbf{red boxes}} while transformations specific to the pre-training step are highlighted in \textcolor{onlineBlue}{\textbf{blue boxes}}.}
          \label{fig:online}
      \end{figure*}

\begin{figure*}[!t]
            \centerline{
                \includegraphics[width=1\textwidth]{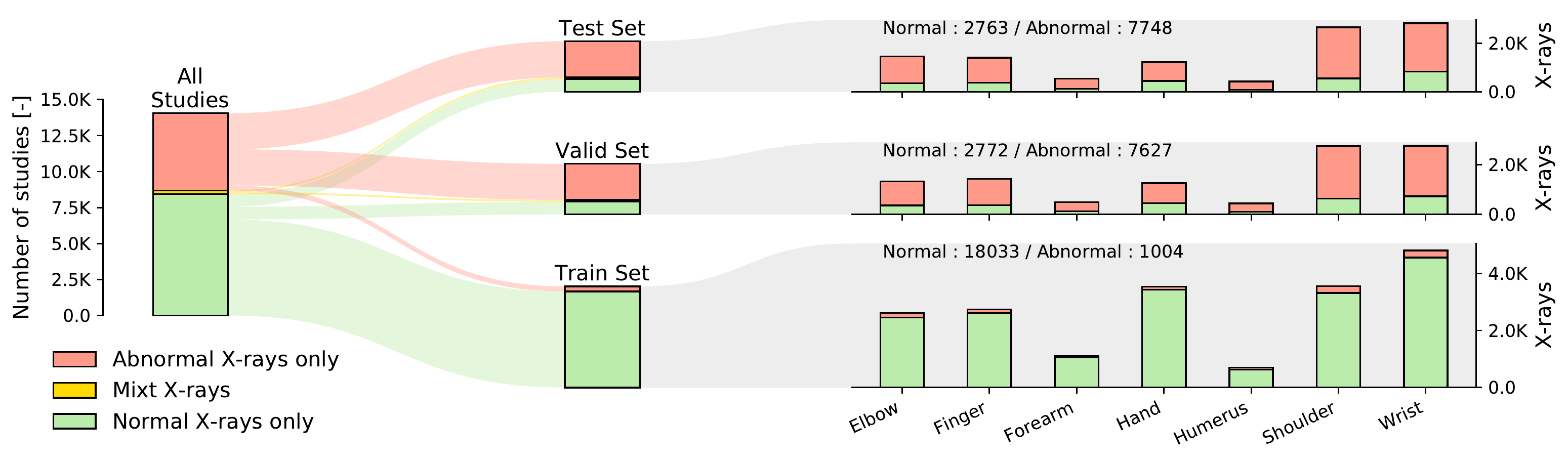}
            }
            \caption{Data splitting summary. This flowchart visually presents the semi-supervised splitting at the study level (\emph{i.e.} patient's body part). Half of the data composed the train set (95\% of normal samples and 5\% of abnormal ones). The remaining data is equally split between the validation and test sets. The bar plots on the left present the distribution of X-ray images in each set's studies for each body part and colored by labels.}
            \label{fig:split}
        \end{figure*}

    We evaluate the proposed method on the Stanford Musculoskeletal Radiograph (MURA) dataset \cite{rajpurkar_mura_2018} composed of around 40,000 upper limb X-ray scans annotated by radiologists as either normal or abnormal. Each bone X-ray is a grayscale image with 512 pixels major axis. The images are first pre-processed offline to extract the X-ray image carrier and generate a body part segmentation mask. Examples of X-ray image carrier detection and segmentation masks are presented in Fig. \ref{fig:rect_crop} and Fig. \ref{fig:segmentation}, respectively. Upon image loading, further online processing is applied to transform the data using heavy augmentations, including random cropping and random color distortions during the pre-training step. Please refer to Fig. \ref{fig:online} for online processing details. The dataset is then split into train, validation, and test sets. Since there are multiple images of the same body part per patient, the dataset is split at the level of the patient’s body part: all images of a body part of a given patient are placed in the same set. This study focuses on semi-supervised training, which implies that we have access mainly to unknown samples and only a few labeled ones. We simulated such a settings with our splitting strategy. In total, we end up with 19,037 training images (95\% of normal samples, with 5\% abnormal samples), 9,910 validation images, and 10,005 test images. A visual summary of the splitting strategy is provided in Fig. \ref{fig:split}.
    
    \subsection{Experimental Setup}
    \label{ssec:exp_setup}
    For all experiments except supervised baseline, we use a ResNet-18 backbone until the average-pooling layer followed by a 2-hidden layers MLP (512$\rightarrow$256$\rightarrow$128) as encoders $\phi_C$ and $\phi_{MAD}$. Besides, to ensure a fair comparison, we apply the same pre-processing for all the baselines. All gradient descent steps are performed using Adam optimizer with the default parameters and a $L_2$ regularization weight of $1e-6$. Each model is trained four times. The models' classification performances are reported through the area under the receiver operating curve (AUC) computed on the anomaly scores. The AE and our pre-training takes 100 epochs with a learning rate of respectively 1e-4 and 1e-3 (decayed by 90\% after 70 and 90 epochs) and a batch size of 16 and 24. For other baselines using AE, we use a mirrored encoder for image reconstruction. We set the hyper-parameter $\eta$ to $1.0$ and $\gamma$ to $0.05$ with $100$ initial hyper-spheres. Our code is available on GitHub.\footnote{\url{https://github.com/antoine-spahr/SELF-TAUGHT-SEMI-SUPERVISED-ANOMALY-DETECTION}}

\section{Results and Discussion}
\label{sec:results}

    To establish competing methods, we compare our method with the state-of-the-art unsupervised AD \cite{tang_abnormal_2019}, semi-supervised AD \cite{ruff_deep_2019} and self-supervised \cite{he_momentum_2020} based AD, with the same data splitting, pre-processing, and encoder bottleneck as ours. Besides, we compare it against DROCC \cite{goyal_drocc_2020} (Sec. \ref{subsec:drocc} in the \textbf{Appendix}) that trains a classifier to distinguish the training samples from their perturbations generated adversarially. Variants of \cite{ruff_deep_2019} are also explored with either joint training or an alternative anomaly score based on the distance to the subspace of normal samples (Sec. \ref{subsec:dsad_joint} and Sec. \ref{subsec:dsad_subspace} in the \textbf{Appendix}). In addition, we train the supervised baseline, ResNet-50, on the binary cross-entropy loss. Table \ref{tab:AUC_results} presents the mean AUC scores for all respective baselines. We observed that our method achieves the best results amongst all the compared methods on the MURA dataset. For example, our method surpasses the best competing method \cite{ruff_deep_2019} by a margin (i.e., 2.89\% in mean AUC). This indicates that our self-supervised method generalizes better against arbitrary imbalanced anomalies. More details about these baselines are provided in the \textbf{Appendix}.

    \begin{table}[!t]
        \centering
        \caption{Evaluation of AUC results for the MURA test set as the mean, and 95\% confidence interval over the N replicates. The middle row shows results with various anomalous ratios.
        }
        \resizebox{\columnwidth}{!}{%
        \begin{tabular}{@{}lcccc@{}}
            \toprule
                                                           & Validation AUC               & Test AUC                     & N            \\ \midrule
            \textbf{Ours}                                  & $\mathbf{77.76 \pm 0.64 \%}$ & $\mathbf{78.04 \pm 0.39 \%}$ & 4            \\
            \textbf{Ours (uni-modal)}                      & $\mathbf{77.55 \pm 0.22 \%}$ & $\mathbf{77.70 \pm 0.27 \%}$ & 4            \\
            Ours (using pretext task from \cite{khosla2020supervised}) & $77.20 \pm 0.89\%$           & $77.61 \pm 0.66\%$           & 4            \\
            Ruff \emph{et al.} \cite{ruff_deep_2019}       & $75.35 \pm 1.04\%$           & $75.15 \pm 0.80\%$           & 4            \\
            Ruff \emph{et al.} (Multi-Modal) AE            & $75.48 \pm 1.39\%$           & $75.16 \pm 1.74\%$           & 4            \\
            Ruff \emph{et al.} (Multi-Modal) Image-Net     & $49.09 \pm 2.30\%$           & $49.13 \pm 3.91\%$           & 4            \\
            Ruff \emph{et al.} (Joint training)            & $74.70 \pm 0.35\%$           & $74.39 \pm 0.34\%$           & 4            \\
            Ruff \emph{et al.} (Joint training Subspace)   & $54.87 \pm 3.08\%$           & $53.68 \pm 5.38\%$           & 4            \\
            He \emph{et al.} \cite{he_momentum_2020}       & $74.94 \pm 0.25\%$           & $74.85 \pm 0.90\%$           & 4            \\
            Tang \emph{et al.} \cite{tang_abnormal_2019}   & $70.86 \pm 1.16\%$           & $71.14 \pm 1.05\%$           & 4            \\
            Goyal \emph{et al.} \cite{goyal_drocc_2020}    & $51.66 \pm 3.02\%$           & $51.49 \pm 6.88\%$           & 2            \\
            Supervised ResNet-50                           & $70.38 \pm 4.18\%$           & $70.17 \pm 4.23\%$           & 4            \\ \midrule
            \textbf{Ours 2.5\%}                            & $\mathbf{74.20 \pm 0.96\%}$  & $\mathbf{74.87 \pm 0.77\%}$  & 2            \\ 
            Ruff \emph{et al.} \cite{ruff_deep_2019} 2.5\% & $72.23 \pm 1.86\%$           & $72.10 \pm 1.78\%$           & 2            \\ \midrule
            \textbf{Ours 10\%}                             & $\mathbf{78.73 \pm 0.03\%}$  & $\mathbf{79.78 \pm 0.49\%}$  & 2            \\ 
            Ruff \emph{et al.} \cite{ruff_deep_2019} 10\%  & $77.66 \pm 0.00\%$           & $78.91 \pm 0.08\%$           & 2            \\ \midrule
            Our pre-training 100-NN                        & $49.38 \pm 0.39\%$           & $50.89 \pm 0.47\%$           & 4            \\ 
            AE 100-NN                                      & $48.96 \pm 1.08\%$           & $49.73 \pm 1.49\%$           & 4            \\ \bottomrule
        \end{tabular}
        }
        \label{tab:AUC_results}
    \end{table}

    \begin{figure}[t]
        \centering
        \centerline{\includegraphics[width=8.5cm]{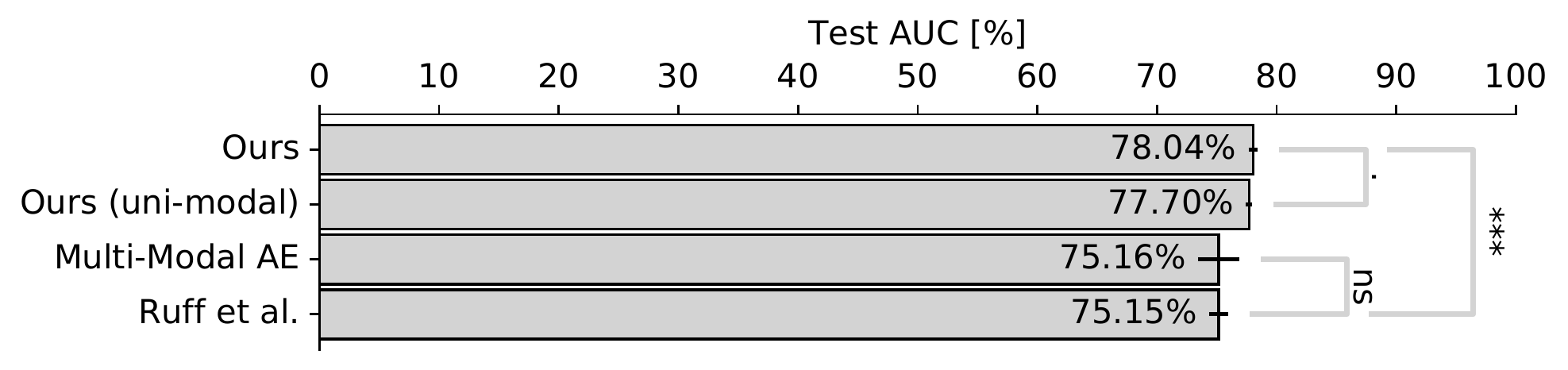}}
        \caption{Test AUC performances presented as $mean\pm 1.96std$. Samples are statistically tested for \emph{H\textsubscript{0}:means are similar}, using a bilateral Welch t-test. The pairs tested are displayed by a gray link with the significance code. \emph{ns} non-significant, . $p_{value} \in [1, 0.1]$, * $p_{value} \in [0.1, 0.05]$, ** $p_{value} \in [0.05, 0.01]$, *** $p_{value} \in [0.01, 0.001]$.}
        \label{fig:results}
    \end{figure}

    \begin{figure}[t]
        \centering
        \centerline{\includegraphics[width=8.5cm]{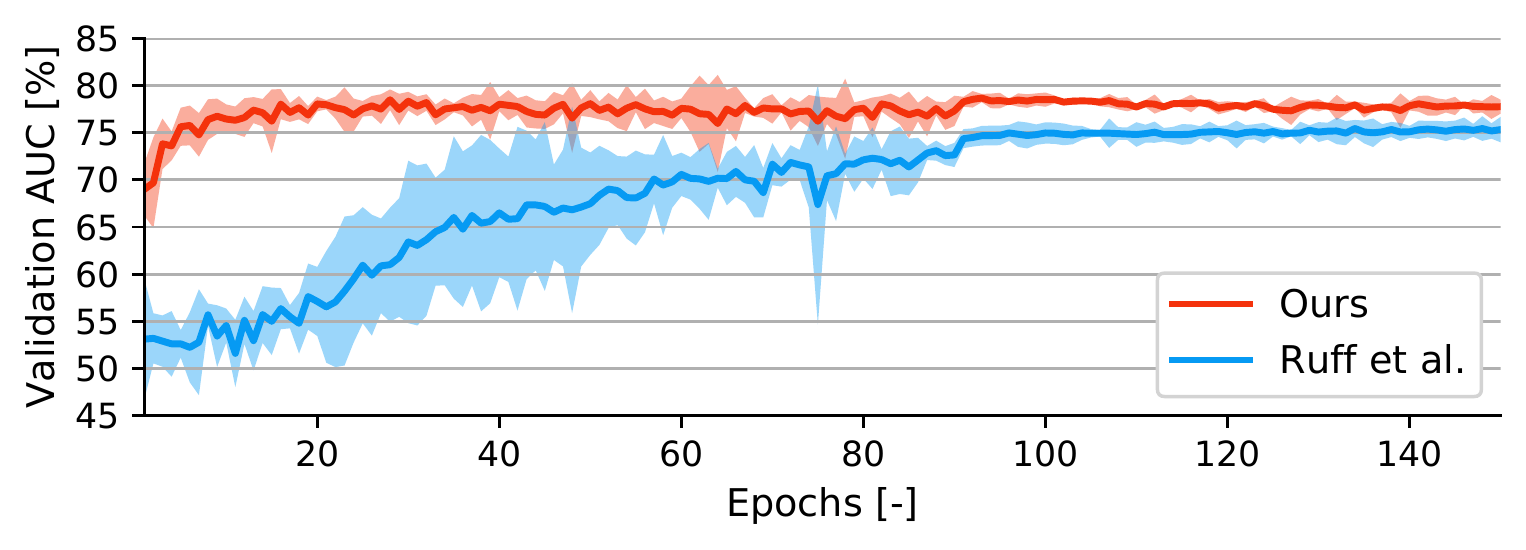}}
        \caption{The comparison of our pre-training with AE pre-training based on validation AUC scores evolving during training. The results are presented with the mean and 95\% confidence interval ($mean\pm 1.96std$) at each epoch.}
        \label{fig:AUC_evolv}
    \end{figure}
    
    \begin{figure}[t]
        \centering
        \centerline{\includegraphics[width=8.5cm]{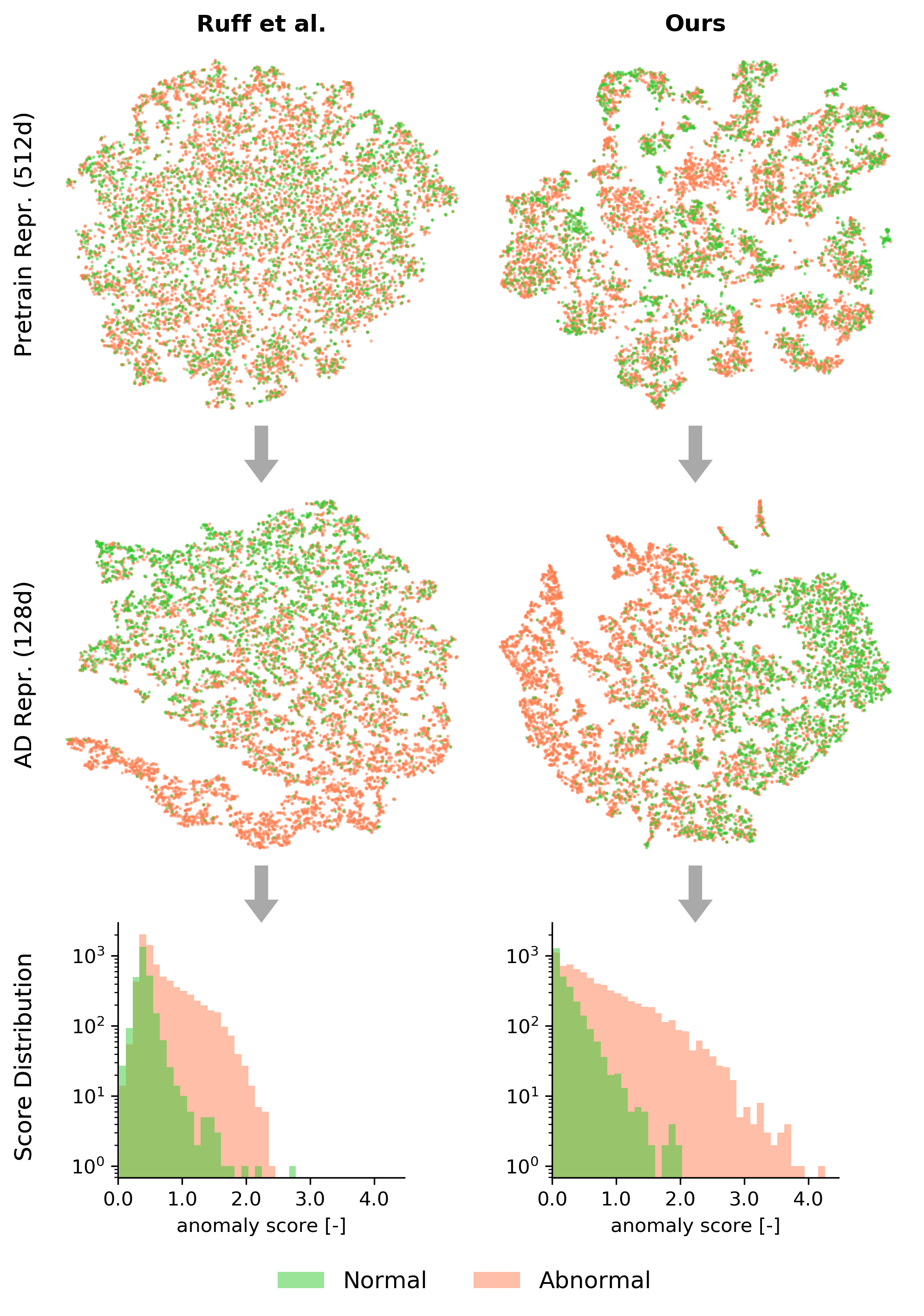}}
        \caption{The t-SNE visualization of the representation learned with the pretext task (top row) and of the final representation learned with the semi-supervised task (middle row) with the resulting anomaly score distribution (bottom row).}
        \label{fig:tsne}
    \end{figure}

    
    \subsection{Ablation Study}
    
    \textbf{Pretext Task.}
    The ablation study is conducted to analyze the pre-training of the proposed method. To do so, we show the 512-dimensional feature representations of the validation set learned by our pre-training, and AE pre-training transformed using t-SNE (Fig. \ref{fig:tsne}, top row). Interestingly, the visualization shows that the representation learned using our pretext task tends to already group samples into normal and abnormal clusters. The network has thus learned generalizable features that are easily tailored to the downstream anomaly detection task. Oppositely, the AE objective tends to ignore the underlying structure of anomalies and thus learns representations fragile to segregate normal and abnormal samples well. After pre-training on the unlabeled images, we advocate meaningful representation of the data obtained by pretext task and fine-tune the network with only a few labeled images to achieve good performance for the anomaly detection task. 
    
    \textbf{Downstream Task.}
    We also conduct ablation studies to investigate the downstream AD task's effect and our method's sensitivity w.r.t the ratio of known anomalous samples. The anomaly score is computed as the mean distance between a sample and the 100 nearest normal training samples in the embedded space ($\mathbb{R}^d$) obtained from the pre-training stage. We chose this optimum number of nearest neighbors experimentally. Anomalous samples should yield a high mean distance compared to normal samples. Table \ref{tab:AUC_results} (bottom) presents the anomaly detection results using our approach and AE representation. The results indicate the AD task is of paramount importance for achieving good performance. Besides, AD results for various anomalous ratios are presented (Table \ref{tab:AUC_results} middle).
    \textbf{Network Convergence.}
    Our method's generalizable features transpose in better performances and a faster optimization of the anomaly detection network.  Indeed, as highlighted in Fig. \ref{fig:AUC_evolv}, the validation AUC after the first epoch of optimization has already raised to $68.98\pm \ 2.77\%$, and then it rapidly reaches the plateau of 78\% in just a few epochs. On the other hand, with an AE initialization, the AUC after one epoch is not even above random ($53.11\pm \ 6.37\%$), followed by a slower increase to reach a plateau at around 75\% after 100 epochs. Thus, the anomaly detection task's convergence is faster when using the proposed pre-training, implying less computational power needed. We further validate our learned feature representation for computing anomaly score using the t-SNE of the final representation in Fig. \ref{fig:tsne} (middle row). Unlike the sparse representation of samples with an AE pre-training, our method could generate more compact normal samples' representations while repulsing abnormal ones. Our method's more proficient training further transposes in a better anomaly score distribution (Fig. \ref{fig:tsne} bottom row). Moreover, unlike the uni-modal framework, our method makes less assumption about the underlying data distribution and can be better optimized to the best suited center(s) (Fig. \ref{fig:results}).

\section{Conclusion}
\label{sec:conclusion}
We presented a new universal method for anomaly detection on musculoskeletal radiographs, which can generalize well, reject unseen distributions, and leverage unlabeled data. Using the semi-supervised multi-modes anomaly detection, we formulated a complex nature of X-rays. Our method outperforms other weakly-supervised and self-supervised anomaly detection baselines in terms of AUC accuracy while being computationally much lighter. Furthermore, ablation studies have shown that the proposed method yields faster convergence than other methods, which proves its ability to learn meaningful and generalized features.




\bibliographystyle{IEEEbib}
\bibliography{references}

\section{Appendix}
\label{sec:appendix}
Here, we briefly present other semi-supervised baselines methods on the MURA dataset with the same preprocessing described in the paper to further validate our proposed contrastive learning-based method. We explored two extensions of deep semi-supervised anomaly detection (DSAD) \cite{ruff_deep_2019} (a joint training of the AE and the hyper-sphere, an alternative distance metric for the anomaly score) and an adversarially robust method, DROCC. 
    
    \begin{figure*}[htbp!]
      \centering

      \begin{subfigure}[b]{1\columnwidth}
        \centerline{\includegraphics[width=\columnwidth]{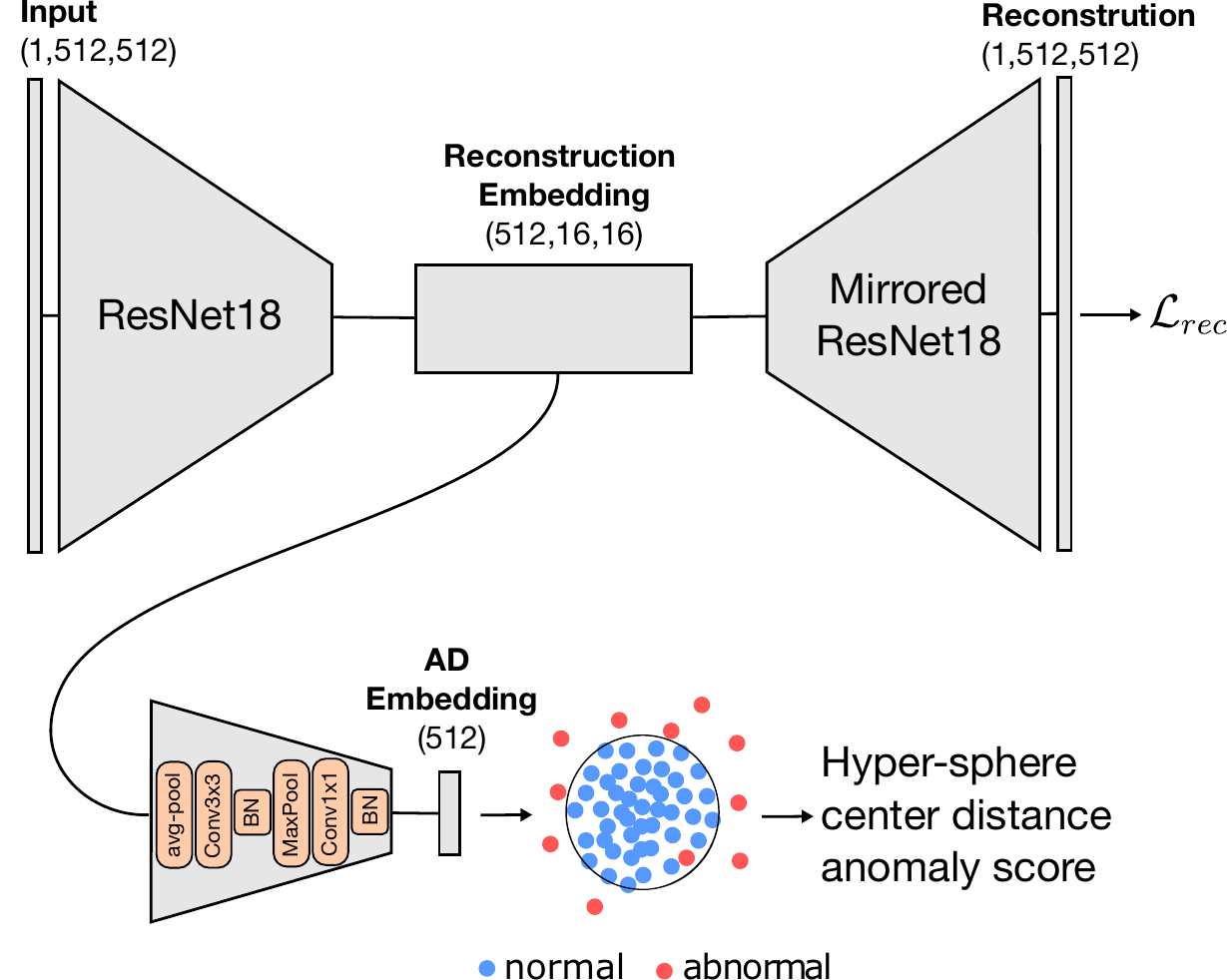}}
        \caption{Joint Training DSAD.}
        \label{fig:jointDSAD}
      \end{subfigure}
      \begin{subfigure}[b]{1\columnwidth}
        \centerline{\includegraphics[width=\columnwidth]{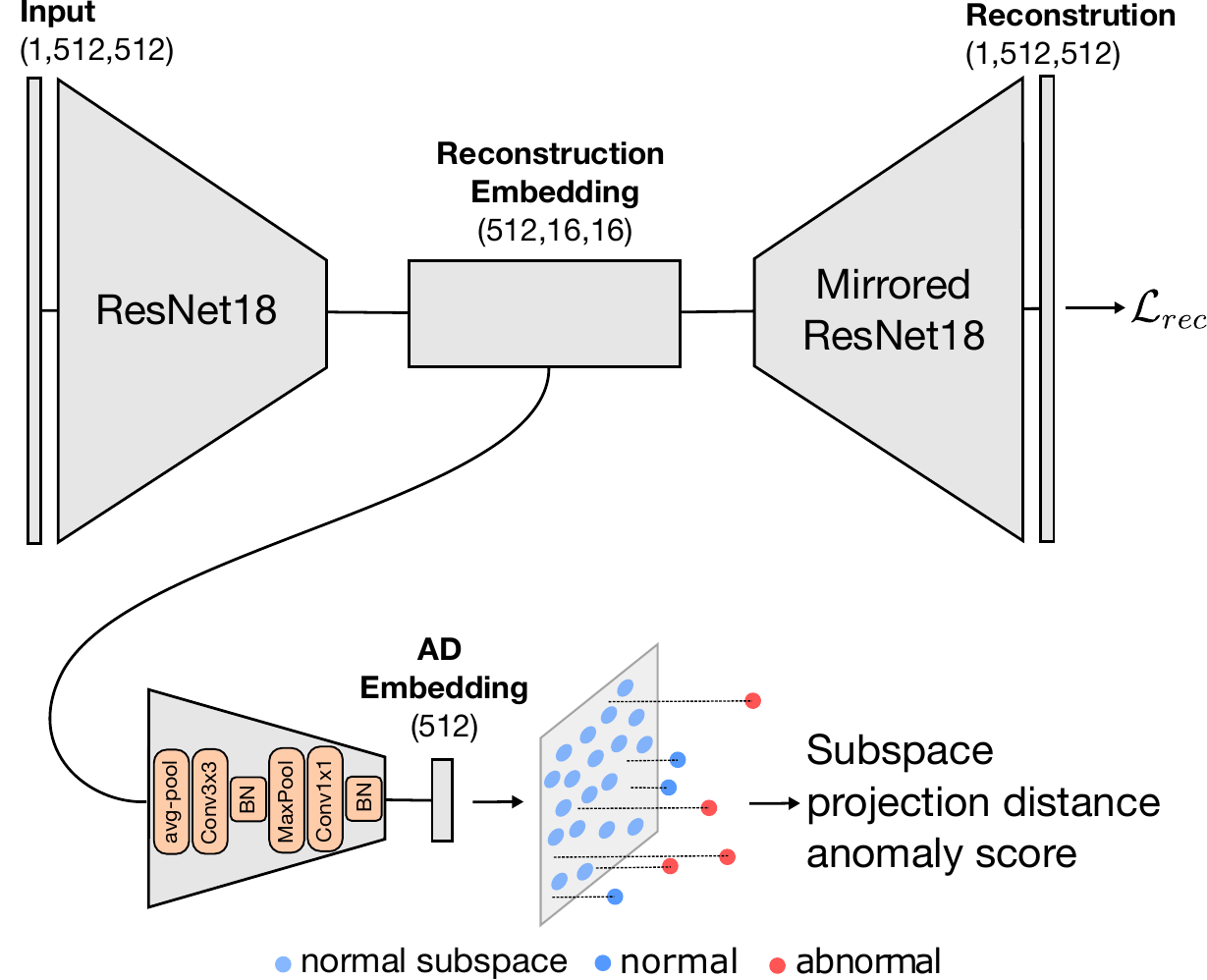}}
        \caption{Joint Training Subspace DSAD.}
        \label{fig:jointSubspaceDSAD}
      \end{subfigure}
      \newline
      \begin{subfigure}{0.8\columnwidth}
        \centerline{\includegraphics[width=\columnwidth]{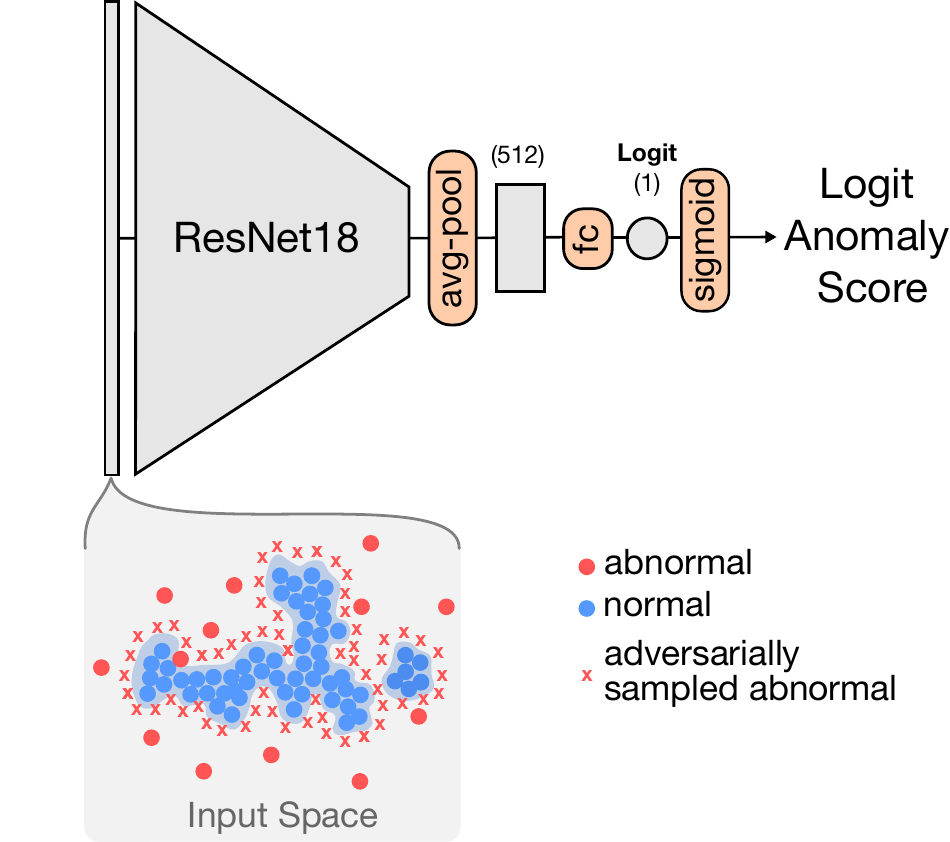}}
        \caption{DROCC}
        \label{fig:DROCC}
      \end{subfigure}

      \caption{Other baselines architectures. (a) DSAD with joint training of the hyper-sphere objective and the AE reconstruction; (b) DSAD with joint training and an alternative subspace distance metric for the anomaly score; (c) DROCC, a semi-supervised method.}
    \end{figure*}
    
    \subsection{Joint Training DSAD}
    \label{subsec:dsad_joint}
        Joint DSAD is an extension of DSAD described in \cite{ruff_deep_2019} in which the separate training of the AE and the hyper-sphere concentration have been merged to be trained together. The AE reconstruction embedding is different from the one generated for anomaly detection purposes to provide proper parallel optimization. The AE loss should act as a regularization of the shared weights and enforce them to stay meaningful for anomaly detection. Joint training also reduces the computation time compared to the original approach. Fig. \ref{fig:jointDSAD} present a scheme of the method.
        
    \subsection{Joint Training Subspace DSAD}
    \label{subsec:dsad_subspace}
        The second extension of DSAD is explored by using a different anomaly score distance. We used the same joint training procedure as described above, but we used the projection distance to the subspace of normal train samples as an anomaly score. The projection distance is computed similarly as in SubspaceNet for Few-Shot learning \cite{devos_subspace_2019}. The projection matrix is computed with 10,000 normal train samples and should represent the normal sample distribution. Afterward, the network's weights are optimized to reduce the distance of normal samples to this subspace and push abnormal samples away. Fig. \ref{fig:jointSubspaceDSAD} present a scheme of the method.
        
    \subsection{DROCC}
    \label{subsec:drocc}
        Deep robust one-class classification \cite{goyal_drocc_2020} is an anomaly detection method that makes use of adversarial robustness in training. The method builds upon the semi-supervised setup with a classifier with a single neuron as output. This neuron value (logit) is used as an anomaly score. The network is trained to yield high value for anomalous samples and low value for normal ones. However, since there are no/few abnormal samples for supervised training, DROCC builds additional abnormal samples by adversarial search close (but not too close) to the normal training samples. The resulting abnormal samples should then drive the network to learn the manifold of normal samples and detect out-of-distribution samples. Fig. \ref{fig:DROCC} presents a scheme of the method.
    
\end{document}